\documentclass[runningheads]{llncs}
\usepackage{graphicx}
\usepackage{array}
\usepackage{subfigure}
\usepackage{setspace}
\usepackage{mathrsfs}
\usepackage{amsmath}
\usepackage{bm}
\usepackage{amsfonts}
\usepackage{ulem}
\usepackage{booktabs}
\usepackage[misc,geometry]{ifsym} 

\usepackage[colorlinks,
            linkcolor=green,
            anchorcolor=red,
            citecolor=blue
            ]{hyperref}

\makeatletter
\newcommand{\printfnsymbol}[1]{\textsuperscript{\@fnsymbol{#1}}}
\makeatother
\newcommand{\PreserveBackslash}[1]{\let\temp=\\#1\let\\=\temp}
\newcolumntype{C}[1]{>{\PreserveBackslash\centering}p{#1}}

\begin{document}

\title{Difficulty-aware Glaucoma Classification with Multi-Rater Consensus Modeling}
\titlerunning{Difficulty-aware Glaucoma Class. with Multi-Rater Modeling}

\author{Shuang Yu \thanks{First two authors contributed equally.} \inst{1} \textsuperscript{(\Letter)}  \and
Hong-Yu Zhou \printfnsymbol{1} \inst{1}  \and
Kai Ma \inst{1} \and
Cheng Bian \inst{1} \and
Chunyan Chu \inst{1} \and
Hanruo Liu \inst{2} \and
Yefeng Zheng \inst{1}
}

\authorrunning{S. Yu et al.}

\institute{
	Tencent Healthcare, Tencent, Shenzhen, China \\
	\email{shirlyyu, hongyuzhou, kylekma, tronbian, yefengzheng@tencent.com} \\
	\and
	Beijing Tongren Hospital, Capital Medical University, Beijing, China \\
}
\maketitle              
%

\begin{abstract}

Medical images are generally labeled by multiple experts before the final ground-truth labels are determined. Consensus or disagreement among experts regarding individual images reflects the gradeability and difficulty levels of the image. However, when being used for model training, only the final ground-truth label is utilized, while the critical information contained in the raw multi-rater gradings regarding the image being an easy/hard case is discarded. In this paper, we aim to take advantage of the raw multi-rater gradings to improve the deep learning model performance for the glaucoma classification task. Specifically, a multi-branch model structure is proposed to predict the most sensitive, most specifical and a balanced fused result for the input images. In order to encourage the sensitivity branch and specificity branch to generate consistent results for consensus labels and opposite results for disagreement labels, a consensus loss is proposed to constrain the output of the two branches. Meanwhile, the consistency/inconsistency between the prediction results of the two branches implies the image being an easy/hard case, which is further utilized to encourage the balanced fusion branch to concentrate more on the hard cases. Compared with models trained only with the final ground-truth labels, the proposed method using multi-rater consensus information has achieved superior performance, and it is also able to estimate the difficulty levels of individual input images when making the prediction.

\keywords{Multi Rater \and Retinal Imaging \and Glaucoma Classification \and Uncertainty Estimation}
\end{abstract}
%
%
\section{Introduction}

Glaucoma is the leading cause of irreversible vision loss world widely and is projected to affect around 111 million people by year 2040 \cite{tham2014global}.
Therefore, the screening and treatment of glaucoma in the early stage play an important role to prevent vision loss. Recent years, there has been an increasing trend in the automatic classification of glaucoma with deep learning methods, including \cite{li2018efficacy,phene2019deep,liu2019development}. However, the reference standard and guideline for glaucoma diagnosis are often not well defined and may vary from one center to another, which might result in disagreement among graders and negatively affect the grading procedure \cite{hammel2019study,phene2019deep}. It is reported that the sensitivity of individual graders for glaucoma ranged from 29.2\% to 73.9\%, with specificity ranged from 75.8\% to 92.6\% \cite{phene2019deep}. Therefore, the inter-rater variability problem constitutes a major impact for the grading procedure of glaucoma.

Glaucoma images, in fact medical images in general, are usually labeled by multiple experts independently, so as to avoid the subjective bias or potential labeling noise of each rater resulted by different levels of expertise, negligence of subtle symptoms, quality of images, etc \cite{schaekermann2019understanding}. 
The final ground-truth label then can be obtained by fusing individual labels using majority vote, average or other fusion strategies \cite{schaekermann2019understanding}.
However, at model training stage, only the final ground-truth label is utilized to train the model and those intermediate labels generated by individual raters are neglected, which contain important information regarding the gradeability or difficulty levels of the images. 

Recently, there have been emerging research works paying attention to the multi-rater labels and inter-rater variability. Alain \textit{et al.} \cite{jungo2018effect} studied the effect of common label fusion techniques on the uncertainty estimation of segmentation tasks. It was observed that the models trained with fused `ground truth' label tended to under-estimate the uncertainty, meanwhile uncertainty generated by models trained with individual labels was able to reflect the underlying expert disagreement \cite{jungo2018effect}. Similar influence of the fused final label was observed by Jensen \textit{et al.} \cite{jensen2019improving} for the skin disease classification task as well, which reported that the classification model trained with fused final label would be over-confident, while the model trained with the label sampling method using inter-rater variability was better calibrated. To better utilize the individual ratings, Guan \textit{et al.} \cite{guan2018said} proposed to predict the labels of each rater individually and then learn the respective weight to make the final prediction. In another similar work, Sudre \textit{et al.} \cite{sudre2019let} proposed to model the individual raters' performance together with their consensus status, which achieved better performance compared with training using the fused final label.

Although those recent studies achieved better performance, the critical information contained in the raw multi-rater gradings regarding the image being an easy/hard case is usually neglected or discarded during the training procedure. It is observed that images with consensus labels generally tend to be easy cases while disagreement gradings tend to be hard or highly uncertain ones, as the labeling consensus among individual graders is highly correlated to the grade-ability and difficulty levels of the images being graded \cite{paletz2016uncovering,schaekermann2019understanding}. Therefore, we believe that the model performance can be further boosted by utilizing the multi-rater agreement/disagreement information.

This research aims to fill the performance gap by leveraging the multi-rater consensus information for the glaucoma classification task. Instead of predicting the labels from individual raters, we propose to use a multi-branch structure to generate three predictions under different sensitivity settings, one with the best sensitivity, one with the best specificity, and one in-between, respectively. It also fulfills the clinical requirement of different sensitivity levels for various application scenarios. In addition, a consensus loss is proposed to encourage the sensitivity branch and specificity branch to generate consistent predictions for images with consensus labels and contradictory predictions for images with disagreement labels. Moreover, cosine similarity between the predictions of the two branches contains important uncertainty information and serves as an indicator of the difficulty level for the input image, which is further utilized to encourage the model to focus more on the hard cases and improve the model performance.


\section{Method}

\begin{figure}[t]
	\centering
	\includegraphics[height=5.8cm, width=11.5cm]{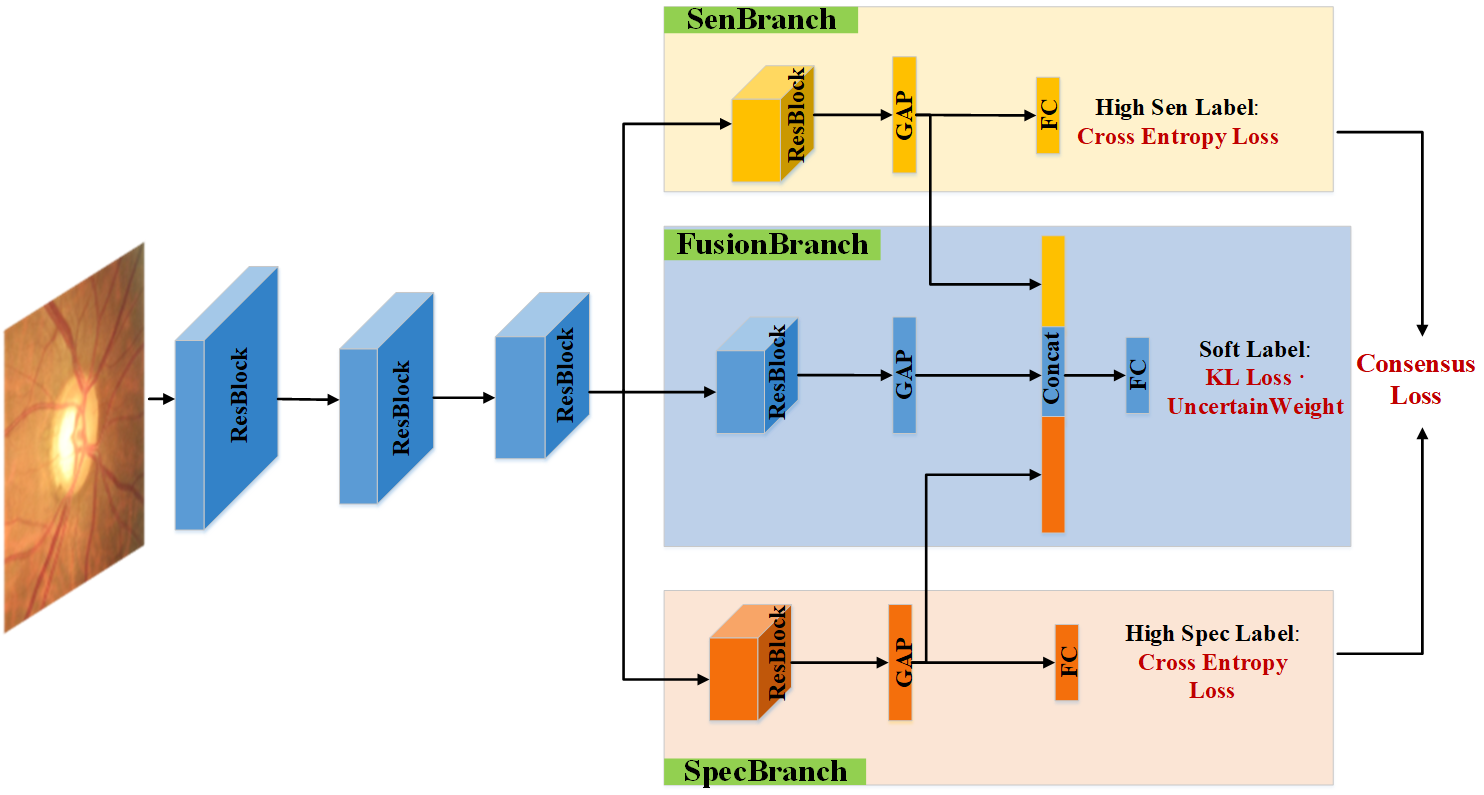}
	\caption{Framework of the proposed system.}
	 \label{Framework}
\end{figure}

Fig.~\ref{Framework} shows the framework of the proposed system, which consists of three branches, corresponding to three different levels of sensitivity and specificity settings, including the sensitivity branch (SenBranch), the specificity branch (SpecBrach) and the balanced fusion branch (FusionBranch), respectively. The three branches share the same weights for the first three ResNet blocks (ResBlock), with ResNet18 \cite{he2016deep} being adopted as the backbone. And then, each branch contains a ResNet block, global average pooling (GAP) and fully connected layers (FC). The extracted GAP features from the SenBranch and SpecBranch are concatenated together with that from the FusionBranch and then fed to the FC layer for the final glaucoma prediction.

\subsection{Consensus Loss}

In order to take advantage of the agreement/disagreement among individual raters, a consensus loss is proposed to encourage the SenBranch and SpecBranch to generate consistent predictions for images with agreement labels and contradictory predictions for images with disagreement labels. The consensus loss is similar to that of the contrastive loss \cite{hadsell2006dimensionality}, generally used in the Siamese network:
\begin{equation}
\begin{aligned}
\label{eq_conloss}
\mathcal{L}_{con}(y'_{sen}, y'_{sp}, a) = & \frac{1}{2}a\left\| y'_{sen} -y'_{sp} \right\| _2^2  + \\& \frac{1}{2}(1-a)\left\lbrace max \left(  0, m - \left\|y'_{sen} -y'_{sp}\right\|_2\right)  \right\rbrace ^2,
\end{aligned}
\end{equation}
where $a$ denotes the consensus label among experts, 1 for consensus and 0 for non-consensus; $ y'_{sen} $ and $y'_{sp} $ represent the model output for the SenBranch and SpecBranch, respectively; $m$ denotes the margin and is set to 1 by default.

\subsection{Uncertainty Estimation}

The prediction consensus between the SenBranch and SpecBranch indicates the difficulty level, i.e., uncertainty, of the images. In this paper, cosine similarity is adopted to measure the distance between the predictions of the two branches. Then, the uncertainty of the model prediction can be estimated with:
\begin{equation}
\label{eq_uncert}
u = 0.5(1-Similarity) = 0.5\left(1-\frac{y'_{sen} \cdot y'_{sp}}{\left\| y'_{sen}\right\| \left\|y'_{sp}\right\|}\right) .
\end{equation}

The obtained uncertainty is further utilized to adjust the relative weight of individual samples for the training of the FusionBranch, so as to encourage the model to concentrate more on the difficult samples.

\subsection{Loss Function for Multi-Branch}

For the training of the proposed model, each branch is optimized individually with the same batch of images, but with different corresponding labels and loss functions. The SenBranch and SpecBranch are trained with the most sensitive and most specifical labels for individual images, the labels of which are determined with a random sampling procedure by assigning different probabilities for individual ratings in the labeling pool. For the SenBranch, the glaucoma labels are set with a higher probability than that of the non-glaucoma labels by repeating the glaucoma labels twice in the label pool; vice versa for the SpecBranch. 
Both branches are optimized with cross entropy loss and consensus loss, as in:
\begin{equation}
\label{eq_loss1}
\mathcal{L}_s(y'_s, y_s, a) =  - \sum_{i=1}^{2} {y_{si} log\left( y'_{si} \right)}   +  \alpha \mathcal{L}_{con}(y'_{sen}, y'_{sp}, a),
\end{equation}
where $y_s$ and $y'_s$ denote the glaucoma label and model output for the SenBranch or SpecBranch; $\alpha$ denotes the relative weight of the consensus loss, empirically set as 0.5 in this research.

For the FusionBranch, instead of training on the final ground-truth, it is trained with soft labels generated from individual rater's grading weighted by their respective accuracy: $y = \frac{\sum_{i=1}^m{w_i r_i}}{\sum_{i=1}^m{w_i}}$, where $m$ is the total number of raters for the image; $r_i$ denotes the raw labels by individual raters; $w_i$ is the weight of the corresponding rater, which is determined by the rating accuracy against the ground-truth label. Furthermore, the soft labels used for the FusionBranch training are clipped to the range of (0.01, 0.99), so as to avoid the potential problems of hard labels.

The Kullback–Leibler divergence loss (KL loss) is then adopted to optimize the model parameters of the FusionBranch. In addition, in order to encourage the model to emphasize the difficult and highly uncertain samples, the estimated uncertainty value is utilized to adjust the relative weight of individual samples, as in:
\begin{equation}
\label{eq_loss2}
\mathcal{L}_f(y, y') = KL(y||y')= \frac { \sum_{i=1}^n{\sum_{j=1}^2{ (1+u_i)y_{ij} \left(log y_{ij} - log y'_{ij} \right) }}}   {\sum_{i=1}^n (1+u_i)},
\end{equation}
where $y$ and $y'$ denote the soft label and model output of the FusionBranch, respectively; $n$ is the total number of samples in a training batch; and $u_i$ is the uncertainty weight for the corresponding sample obtained with Eq. \ref{eq_uncert}.


\section{Experimental Results}

A total of 6,318 color fundus images with acceptable image quality were collected from Beijing Tongren Hospital, with approval obtained from the institutional review board of Tongren Hospital. The images were labeled following the adjudication process, with two certified ophthalmologists and one senior glaucoma specialist involved in the grading procedure.
Each image was independently labeled by two certified ophthalmologists in the first stage.
If consensus label was reached, then the grading process was completed. Otherwise, the image would be passed to the senior glaucoma specialist, who had access to the individual ratings of the first stage and graded the image based on his or her expertise. After the adjudication grading procedure, 2,171 images are graded with consensus glaucoma label, 2,315 images with consensus non-glaucoma label, 781 images with non-consensus glaucoma label and 1,051 images with non-consensus non-glaucoma label. At the model training stage, 60\% of the images are randomly selected as the training set, 15\% as the validation set and the rest 25\% are reserved for test purpose.

Apart from the private dataset, two publicly available datasets, REFUGE (test set) \cite{orlando2020refuge} and DRISHTI \cite{sivaswamy2015comprehensive}, are also adopted. The REFUGE test set contains 40 glaucoma images and 360 non-glaucoma images, and the performance of two individual experts who are not part of the ground-truth labeling group is also reported \cite{orlando2020refuge}. Meanwhile, the DRISHTI dataset contains 70 glaucoma images and 31 non-glaucoma images \cite{sivaswamy2015comprehensive}. Each image is independently graded by five experts with majority vote being adopted as the ground truth. Note that both datasets are used for direct model inference without any further training or fine-tuning, so as to verify the generalization capability of the proposed model.

As pathologies of glaucoma concentrate on the optic disc and surrounding regions, the three-disc-diameter region around the disc center is cropped as the region-of-interest (ROI) and resized to the dimension of $256\times256$ pixels before being fed to the network. All experiments are performed on an NVIDIA Tesla P40 GPU with 24 GB of memory. 
The Adam optimizer is adopted to optimize the model with a batch size of 32 and maximum training epochs of 50. The initial learning rate is set as $2\times 10^{-4}$ and halved every 15 epochs. Data augmentation strategies, including random cropping, rotation, horizontal flipping and color jitting, are utilized during the training procedure, so as to increase the diversity of the training data.

\subsection{Ablation Studies}

Comprehensive ablation studies have been performed to evaluate the effectiveness of different modules proposed in this research, including the multi-branch structure (MultiBr), consensus loss (ConLoss) and uncertainty loss (Uncerty). The comparison baseline model shares the same backbone as the proposed model, i.e., ResNet18, and it is trained with the final ground-truth label. Five metrics are adopted to evaluate the model performance, including accuracy (Acc), sensitivity (Sen), specificity (Spec), F1 score (F1) and area under curve (AUC). 

Detailed results of the ablation studies are listed in Table \ref{tab_abstudy}. We have also evaluated the performance of the two graders in the first stage. On the test set, the two experts achieve an F1 score of 86\% and 82.9\%, respectively, indicating the apparent challenges of glaucoma labeling even for certified ophthalmologists. In contrast, the baseline model trained with ground-truth surpasses the performance of the stage one raters, with an F1 score of 88.6\%. By introducing the multi-branch model structure and taking advantage of raw gradings, the F1 score is improved by 1.19\% over the baseline, with an AUC score of 96.93\%. The effectiveness of consensus loss and uncertainty loss is also verified, yielding an F1 improvement of 0.71\% and 1.23\%, respectively. At last, the proposed multi-branch model combining consensus loss and uncertainty loss achieves the best performance with an F1 score of 91.89\% and AUC value of 97.94\%.

\begin{table}[!t]
	\centering
	\caption{The ablation study results of multi-rater consensus model ($\%$).}
	\label{tab_abstudy}
	\begin{tabular}{p{1.4cm}<{\centering}|p{1.4cm}<{\centering}|p{1.4cm}<{\centering}|p{1.2cm}<{\centering}|p{1.2cm}<{\centering}|p{1.2cm}<{\centering}|p{1.2cm}<{\centering}|p{1.2cm}<{\centering}}
		\hline
		\multicolumn{3}{c|}{Methods } & Acc & Sen & Spec &  F1 & AUC  \\ \hline
		\multicolumn{3}{c|}{Expert 1} & 86.71 & 88.11 & 85.50 & 86.03 & - \\ 
		\multicolumn{3}{c|}{Expert 2} & 83.78 & 84.76 & 82.94 & 82.92 & - \\ \hline
		\multicolumn{3}{c|}{Baseline} & 89.47 & 88.12 & 90.64  & 88.60 & 95.83 \\ \hline

		MultiBr &  - & - & 90.53  &   89.60 & 91.35  & 89.79 & 96.93    \\ \hline
		MultiBr & ConLoss  & - & 91.35 & 88.66 & 93.68 &  90.50  & 97.43  \\ \hline
		MultiBr & -  & Uncerty & 91.70 &  90.28 & 92.94 & 91.02  & 97.52  \\ \hline
		MultiBr & ConLoss  & Uncerty  & \textbf{92.54} & \textbf{90.96} & \textbf{93.92} & \textbf{91.89} &  \textbf{97.94} \\ \hline
	\end{tabular}
\end{table}

\subsection{Comparison with Existing Methods}

We have also compared the proposed method with other research works that utilize the multi-rater labels, including the random sampling method (RandLabel) used by Jensen \textit{et al.} \cite{jensen2019improving} and individual rater modeling (IndiRaters) proposed by Guan \textit{et al.} \cite{guan2018said}. As listed in Table \ref{tab_result2}, the comparison is individually performed on the consensus data, non-consensus data and all data combined together. The proposed method achieves the optimal performance across all the three scenarios, especially for the non-consensus data, with a dramatical performance gain of 4.1\% for sensitivity and 4.9\% for specificity over the current best methods for the non-consensus data. Furthermore, all the comparison methods achieve a superior performance on the consensus data, with both sensitivity and specificity close to or above 95\%. However, for the non-consensus data, there is a dramatical drop of model performance, even for the proposed model. This implies that the images with consensus label tend to be easy or typical cases, either typical normal cases or typical glaucoma images in the advanced stage. In contrast, for the images that grading experts hold different opinions, i.e., the non-consensus data, there is a high probability that the images are hard cases or non-typical cases, difficult for both human graders and deep learning models. Moreover, we have also evaluated the performance of the proposed SenBranch and SpecBranch in Table \ref{tab_result2}. Compared with the FusionBranch, the SenBranch achieves a sensitivity of 99.32\%, and the SpecBranch achieves a specificity of 97.89\%, reflecting the effectiveness of the SenBranch/SpecBranch in fulfilling the designated purpose.

%
%
%
%

\begin{table}[!t]
	\centering
	\caption{Comparison with other multi-rater research method ($\%$).}
	\label{tab_result2}
	\scalebox{0.85}{
	\begin{tabular}{p{2.2cm}<{\centering}|p{1.2cm}<{\centering}p{1.2cm}<{\centering}p{1.2cm}<{\centering}|p{1.2cm}<{\centering}p{1.2cm}<{\centering}p{1.2cm}<{\centering}|p{1.2cm}<{\centering}p{1.2cm}<{\centering}p{1.2cm}<{\centering}}
		\hline
		Methods& \multicolumn{3}{c|}{Consensus Data}&\multicolumn{3}{c|}{Non-Consensus Data}&\multicolumn{3}{c}{All Data} \\ \cline{2-10}  
		-&  Sen & Spec & AUC &  Sen & Spec & AUC &  Sen & Spec & AUC \\ \hline
		
		Expert 1 & 100 & 100 & - &  55.06 & 52.99 & - &  88.11 & 85.50 & -  \\ 
		Expert 2 & 100 & 100 & - &  42.38 & 44.72 & - &  84.76 & 82.94 & - \\ \hline
		
		Baseline  & 94.68 & 97.11 & 99.42 & 69.90 & 76.32& 80.10   & 88.12 & 90.64  & 95.83 \\ 
		RandLabel \cite{jensen2019improving} & 94.86 & 98.47 & 99.58 & 53.43 & 69.77  & 64.82  & 83.58 & 89.73 & 95.13 \\ 
		IndiRaters \cite{guan2018said} & 95.78 & 97.79 & 99.63  &  71.94 & 78.57 & 82.71 &  89.47 & 91.81 & 96.81 \\ \hline

		Proposed &  \textbf{96.33} &  \textbf{98.64} & \textbf{99.76}  & \textbf{76.02} &  \textbf{83.46} & \textbf{87.49} & \textbf{90.96}& \textbf{93.92} & \textbf{97.94} \\ \hline
		SenBr &  \textit{99.63} &  85.06 & 99.65  & \textit{98.47} &  34.59 & 86.43 & \textit{99.32} & 69.36 & 97.49 \\ 
		SpecBr &  88.26 &  \textit{100} & 99.68  & 44.90 &  \textit{93.23} & 85.45 & 76.79 & \textit{97.89 }& 97.50 \\ \hline

	\end{tabular}
}

\end{table}

\subsection{Model Performance on Public Datasets}

\begin{figure}[t]
	\centering
	\includegraphics[height=4.8cm]{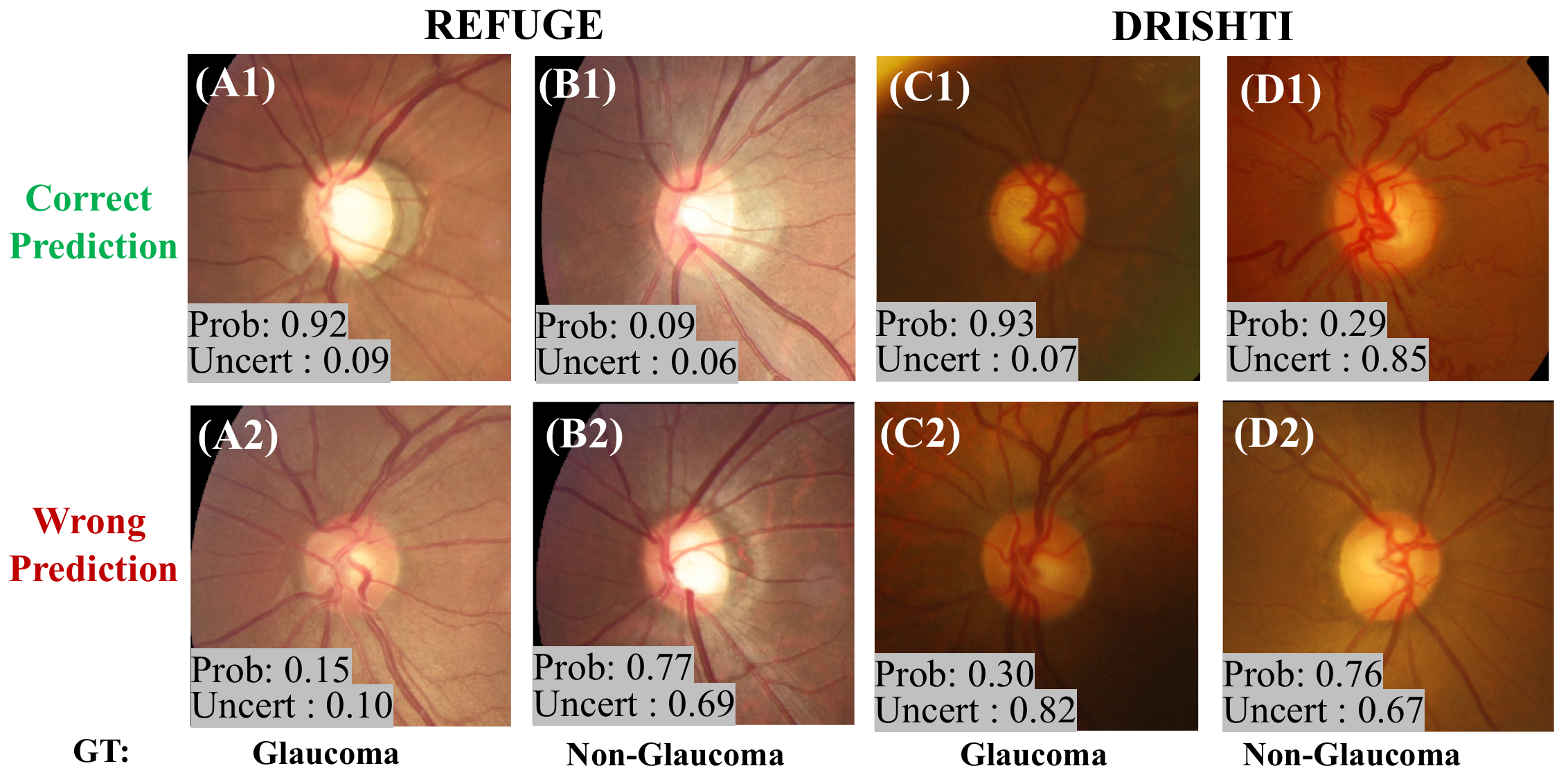}
	\caption{Representative results on the REFUGE and DRISTI datasets. Top row: correct predictions; bottom row: wrong predictions. (A1-2), (B1-2): images from the REFUGE test set; (C1-2), (D1-2): images from the DRISHTI dataset. }
	\label{fig_RepResult}
\end{figure}

In order to verify the generalization capability of the proposed model, we have also tested the model performance on two publicly available datasets, REFUGE and DRISHTI. Note that no further training or fine-tuning is performed on the two datasets. As listed in Table \ref{tab_drishti}, the model achieves the state-of-the-art performance on the DRISHTI dataset, with an AUC improvement of 9.2\% over the current best performance  \cite{diaz2019cnns}. Comparing to the individual expert's grading, the sensitivity and F1 score of the proposed model exceed two out of the five grading experts. As for the REFUGE test dataset (Table \ref{tab_refuge}), the model achieves an AUC value of 96.83\%, which is better than the 3rd place solution on the challenge leaderboard. Concerning that the model is not trained or fine-tuned on the REFUGE data, the result is satisfactory. Especially, when comparing the model performance with two experts, the SenBranch achieves higher sensitivity and specificity than both of the experts.

\begin{table}[!h]
	\centering
	\caption{Performance comparison of glaucoma classification on DRISHTI ($\%$).}
	\label{tab_drishti}
	\begin{tabular}{c|p{1.2cm}<{\centering}|p{1.2cm}<{\centering}|p{1.2cm}<{\centering}|p{1.2cm}<{\centering}|p{1.2cm}<{\centering}}
		\hline
		Methods & Acc & Sen & Spec & F1 & AUC \\
		\hline

		Expert 1 & 91.09 & 95.71 & 80.65 & 93.71 & - \\
		Expert 2 & 87.13 & 81.43 & 100 & 89.76 &- \\
		Expert 3 & 91.09 & 92.86 & 87.10 & 93.53 &- \\
		Expert 4 & 86.14 & 92.86 & 70.97 & 90.28 &- \\
		Expert 5 & 85.15 & 80.00 & 96.77 & 88.19 &- \\ \hline
		
		Sivaswamy\textit{ et al.} (2015) \cite{sivaswamy2015comprehensive}  & - & 81.0 & 72.0& - & 79.0 \\
		Diaz-Pinto\textit{ et al.} (2019) \cite{diaz2019cnns}  & 75.25 & 74.19 & 71.43 & - & 80.41 \\ \hline
		Proposed  & \textbf{86.14} &\textbf{ 91.43} & \textbf{74.19} & \textbf{90.14} & \textbf{89.63}  \\ 
		SenBr  & 78.22 & \textit{98.57} & 32.26 & 86.25 & 87.70 \\ 
		SpecBr  & 74.26 & 68.57 & \textit{87.10} & 78.69 & 84.75  \\ \hline
		
	\end{tabular}

\end{table}

\begin{table}[!h]
	\centering
	\caption{Performance comparison of glaucoma classification on REFUGE ($\%$).}
	\label{tab_refuge}
	\begin{tabular}{c|p{1.5cm}<{\centering}|p{1.5cm}<{\centering}|p{1.5cm}<{\centering}|p{1.5cm}<{\centering}}
		\hline
		Methods & Acc & Sen & Spec & AUC \\
		\hline

		Expert 1 \cite{orlando2020refuge} & 90.50 & 85.00 & 91.11  & - \\
		Expert 2 \cite{orlando2020refuge} & 90.75 & 85.00 & 91.39  & - \\ \hline
		1st Place \cite{orlando2020refuge}  & - & 97.52 & 85.00  & \textbf{98.85}\\ 
		2nd Place \cite{orlando2020refuge}  & - & \textbf{97.60} & 85.00  & 98.17 \\ 
		3rd Place \cite{orlando2020refuge}  & - & 95.00 & 85.00 & 96.44 \\ \hline
		FusionBr  & \textbf{98.0} & 82.50 & \textbf{99.72 } & 96.83 \\ 
		SenBr  & 92.0 & \textit{92.50} & 91.94  & 96.44 \\ 
		SpecBr  & 96.75 & 67.50 & \textit{100} & 93.47 \\ \hline  
	\end{tabular}
\end{table}

Fig. \ref{fig_RepResult} demonstrates several representative results for the correct and wrong predictions on the REFUGE and DRISHTI datasets. In the same time of predicting for the glaucoma probability, the proposed model is also able to estimate the difficulty level, i.e. uncertainty,  via the outputs of SenBranch and SpecBranch. Especially, we have also checked the raw gradings of individual experts for the listed wrong predictions. For images A2 and B2 from REFUGE, the ground-truth labels are glaucoma and non-glaucoma, respectively. However, the two independent ophthalmologists unanimously graded the two images as non-glaucoma and glaucoma, same as the model prediction. For images C2 and D2 from DRISHTI, the uncertainty values estimated by the model are high. When referring to the raw gradings by five experts, 4/5 of the experts labeled C2 as glaucoma and 1/5 as non-glaucoma; 3/5 experts labeled D2 as non-glaucoma and 2/5 as glaucoma, indicating that the glaucoma grading is challenging for experts as well.


\section{Conclusion}

In this paper, we proposed to leverage the multi-rater consensus information contained in the raw expert gradings to enhance the model performance. Ablation studies have validated the effectiveness of the proposed method. It has achieved the state-of-the-art classification performance on the publicly available DRISHTI dataset and satisfactory performance on the REFUGE test set with direct model inference. The proposed model has achieved comparable or better performance than the experts. Future works will continue to explore the potential influence of multi-rater consensus on other deep learning related tasks.

\subsubsection{Acknowledgment}
This work was funded by the Key Area Research and Development Program of Guangdong Province, China (No. 2018B010111001), National Key Research and Development Project (No. 2018YFC2000702) and Science and Technology Program of Shenzhen, China (No. ZDSYS201802021814180).

\footnotesize
\bibliographystyle{splncs04}
\bibliography{paper1015}

\end{document}